\def\year{2019}\relax
\documentclass[letterpaper]{article} 
\usepackage{aaai19}  
\usepackage{times}  
\usepackage{helvet}  
\usepackage{courier}  
\usepackage{url}  
\usepackage{graphicx}  

\usepackage{booktabs} 
\usepackage{graphicx}

\usepackage{amsmath} 
\usepackage{tikz}
\usetikzlibrary{bayesnet}
\usepackage{verbatim}

\begin{document}

\title{Inferring Concept Prerequisite Relations from\\Online Educational Resources}

\author{
Sudeshna Roy\\ 
Videoken, Bangalore, India\\
mail.sudeshna.roy@gmail.com
\And
Meghana Madhyastha \\ 
IIIT-B, India \\
meghana.madhyastha@iiitb.org
\And
Sheril Lawrence\\ 
IIIT-B, India\\
sheril.lawrence@iiitb.org
\And 
Vaibhav Rajan\\
National University of Singapore\\
vaibhav.rajan@nus.edu.sg
}

\maketitle

\begin{abstract}
The Internet has rich and rapidly increasing sources of high quality educational content. Inferring prerequisite relations between educational concepts is required for modern large-scale online educational technology applications such as personalized recommendations and automatic curriculum creation. We present PREREQ, a new supervised learning method for inferring concept prerequisite relations. PREREQ is designed using latent representations of concepts obtained from the Pairwise Latent Dirichlet Allocation model, and a neural network based on the Siamese network architecture. PREREQ can learn unknown concept prerequisites from course prerequisites and labeled concept prerequisite data.
It outperforms state-of-the-art approaches on benchmark datasets and can effectively learn from very less training data.
PREREQ can also use unlabeled video playlists,
a steadily growing source of training data,
to learn concept prerequisites, thus obviating the need for manual annotation of course prerequisites. 

\end{abstract}

\section{Introduction}

A concept $C_1$ is generally called a \emph{prerequisite} to another concept $C_2$ if the knowledge of $C_1$ is necessary to understand $C_2$. Such dependencies are natural in cognitive processes when we learn, organize, and apply knowledge \cite{laurence1999concepts}. Prerequisite relations at a different level -- between courses -- are commonly found in university curricula. Course-level prerequisites have been manually created by experts over decades and often form a guide to prerequisites between the more granular concepts within the courses. For instance, the course Linear Algebra is usually a prerequisite to the course Machine Learning. Several concepts in a course on Linear Algebra are prerequisites to concepts in a course on Machine Learning, e.g. Eigen Analysis is a prerequisite to Principal Components Analysis.

\begin{figure}[ht]
  \begin{center}
	    \includegraphics[width=0.55\textwidth]{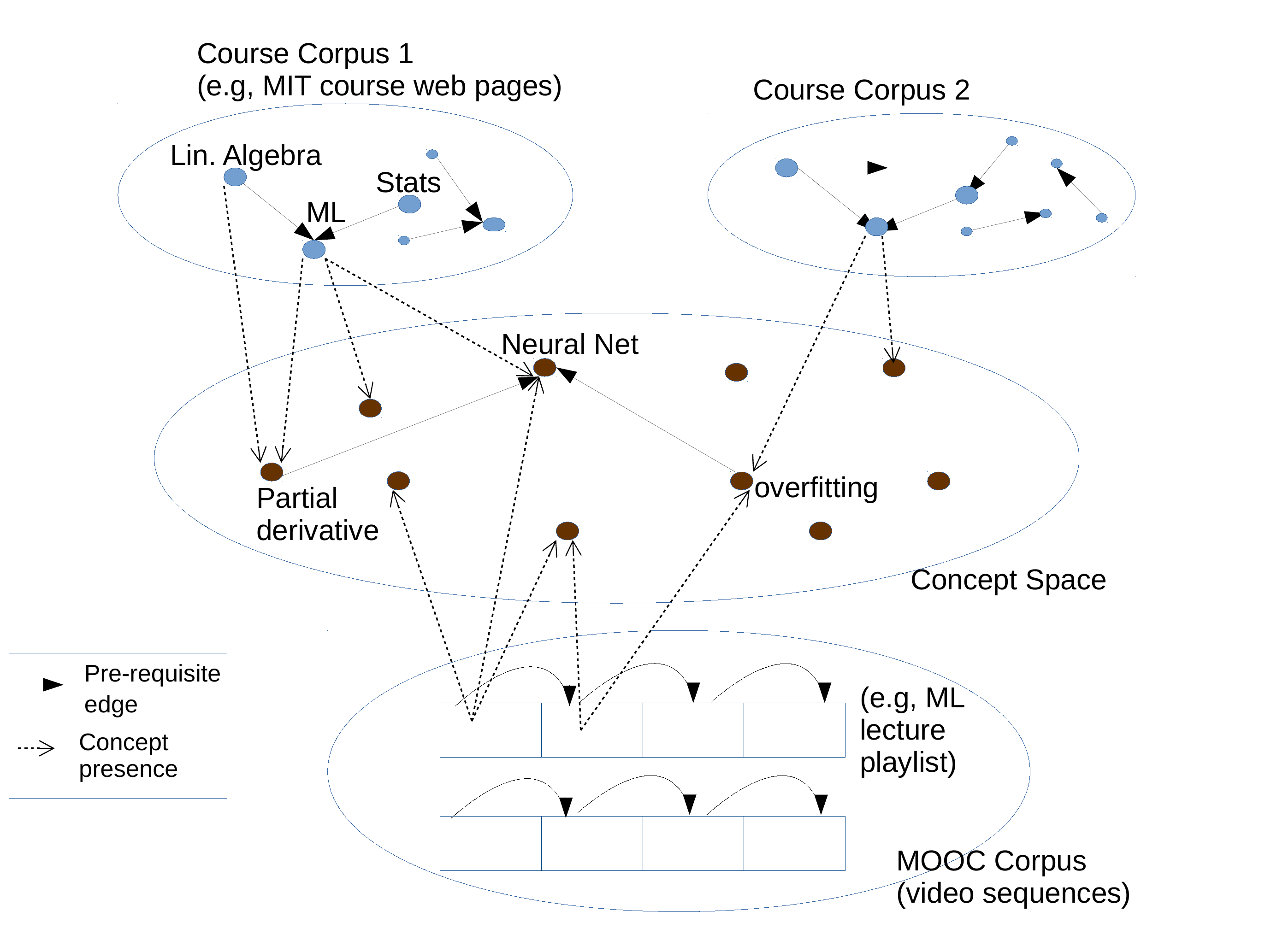}
  \end{center}
  \caption{PREREQ learns unknown prerequisite edges in the concept space, using (1) known concept prerequisite edges and (2) either course prerequisites or video playlists.}
  \label{fig:illustration}
\end{figure}

While textual information about courses has been increasing steadily on the Internet over the years, recently there has been a tremendous growth in online educational data through Massive Open Online Courses (MOOCs) as well as freely accessible videos and blogs from experts. This, in turn, has spurred the development of new applications for personalized online education such as 
automatic reading list generation \cite{jardine2014automatically}, 
automatic curriculum planning \cite{liu2016learning}, and automated evaluation of curricula \cite{rouly2015we}. Concept prerequisite relations play a fundamental role in all these applications.

The value of concept prerequisite maps has been recognized and studied in educational psychology \cite{novak1990concept} 
and these relations were manually obtained by domain experts. 
Such a manual process is not scalable
in modern online applications that aim to (a) serve students from varying educational backgrounds and (b) generalize to any domain. Hence there is a need to develop methods that can automatically infer pair-wise concept prerequisite relations.

Inference of prerequisite relations has been studied in other contexts, e.g. from Wikipedia \cite{Talukdar:2012}, in databases \cite{yosef2011aida} and from text books 
\cite{liang2018investigating}. 
These tools can be leveraged but cannot be directly used to detect prerequisite relations from online resources like MOOCs, due to the complexity and scale of courses and educational concepts involved \cite{acl2017prerequisite}. 
There has been growing interest in designing
algorithms specifically to infer educational concept prerequisites \cite{liu2016learning,eaai17,acl2017prerequisite}.

In this paper, we develop PREREQ, a new supervised learning approach to inferring concept prerequisites. Similar to the problem setting assumed in previous studies, we assume that prerequisites between courses are known and different courses share an underlying concept space. In addition, we assume that some concept prerequisites are also available to train a supervised model. 
Manual annotation of course prerequisites, although available, may be hard to scale.
We show that PREREQ can also effectively learn from unlabeled video playlists, available through MOOCs. 

Figure \ref{fig:illustration} shows a schematic of the underlying concept space shared across different courses from different universities and over different video playlists. We use known course prerequisites or temporal ordering of videos along with labeled training data of concept prerequisites to predict unknown concept prerequisites. Our method uses latent representations of concepts obtained through a Pairwise- Link Latent Dirichlet Allocation (LDA) model \cite{nallapati2008joint}, a model for citations in document corpora. These representations are then used to train a neural network based on the Siamese architecture \cite{siamese} to obtain a binary classifier that can predict, for a given ordered pair of concepts, whether or not a prerequisite relation exists between them.

To summarize, our contributions are:

\begin{itemize}

\item We develop PREREQ, a method to predict unknown concept prerequisites from (1) labeled concept prerequisites and (2) course prerequisite data or video playlists. 
PREREQ uses the pairwise-link LDA model
to obtain vector representations of concepts and a Siamese network to predict unknown concept prerequisites.
\item Our extensive experiments demonstrate the superiority of PREREQ over state-of-the-art methods for inferring prerequisite relations between educational concepts. We also empirically demonstrate  that PREREQ can effectively learn from very less training data.

\end{itemize}

\section{Problem Statement}
Let $C$ be the \emph{concept space}, the set of all concepts of interest, that is assumed to be fixed and known in advance. A concept may be a single word (e.g. ``vector'') or a phrase (e.g. ``machine learning'').
Let $G_C(C,E_C)$ be a directed acyclic graph, called \emph{concept graph}, whose nodes represent concepts and edges represent prerequisite dependency, i.e., $E_C$ contains the directed edge $(c_i, c_j)$ if and only if concept $c_i$ is a prerequisite of concept $c_j$. 

We infer the edges of the concept graph from known prerequisite relations between documents, where documents are text sources containing the concepts of interest.  
Examples of such documents include course web pages with known course prerequisites.
We assume as input a set of text documents $\mathcal{D}$ and a \emph{document graph}, a directed acyclic graph $G_D(\mathcal{D},E_D)$, whose nodes represent documents and edges represent document prerequisite dependency, i.e., $E_D$ contains the directed edge $(d_i, d_j)$ if and only if document $d_i$ is a prerequisite of document $d_j$.
Each document is represented by the concepts contained in it, i.e. $d_i = \{C \cap W_i \}$ where $W_i$ is the set of $n$-grams in document $d_i$, for $n \in \{1,2,3\}$.

For a given set of concepts $C$, we want to infer concept prerequisites $E_C$ from the known document graph $G_D$ and the set of text documents $\mathcal{D}$.
In the supervised setting, some concept prerequisites, denoted by the training set, $E_{CT}$, are known, and the remaining, $E_{CU}$, are unknown, where $E_{CT} \cup E_{CU} = E_C$ and $E_{CT} \cap E_{CU} = \phi$. 
The problem can be stated as, for a given set of concepts $C$, documents $\mathcal{D}$, document graph $G_D(\mathcal{D},E_D)$ and known concept prerequisites $E_{CT}$, predict the unknown concept prerequisites $E_{CU}$.

\section{Our Approach}
\vspace{-0.4cm}
\begin{figure}[!h]
  \begin{center}    \includegraphics[width=0.9\linewidth]{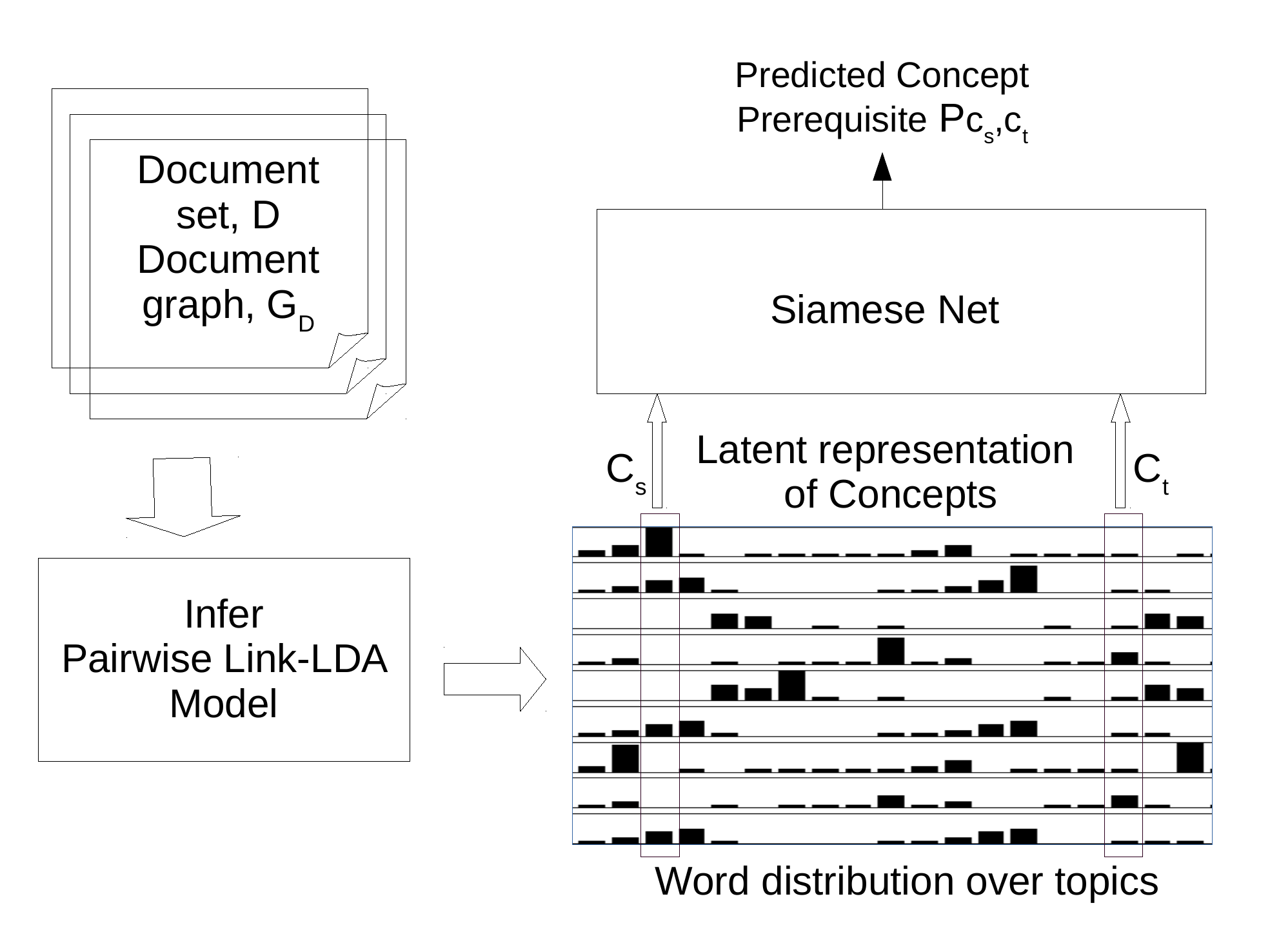}
  \end{center}
  \vspace{-0.6cm}
  \caption{PREREQ Algorithm: latent representations of concepts are obtained using the Pairwise-link LDA model; known concept prerequisite relations are used to train a Siamese network to identify prerequisites.}
  \label{fig:schematic_model}
\end{figure}
Latent topic models for text with citation links between documents have been studied extensively and we study the applicability of one such well-known model, the pairwise-link LDA model~\cite{nallapati2008joint}, to address the problem of concept prerequisite inference.
Our experiments reveal that the latent representations obtained from this model in itself does not have sufficient discriminatory signal. In particular they are a good measure of concept relatedness but not of prerequisite directionality.  
However, learning unsupervised latent representations through a generative probabilistic model helps in disentangling causal factors by discovering underlying causes such as organization of explanatory factors, natural clustering, sparsity and simplicity \cite{bengio2013representation}.
So, we use these latent topic representations of concepts to train a neural network based on the Siamese architecture~\cite{siamese} that can identify prerequisite relations.

A schematic view of our method, called PREREQ, is shown in Figure \ref{fig:schematic_model}. 
The input documents $\mathcal{D}$ and the document graph $G_D(\mathcal{D},E_D)$ are used to learn the pairwise-link LDA model.
Latent representations of the concepts are obtained from this model and used along with known prerequisites $E_{CT}$ to train a Siamese Network.
The following sections describe the details of PREREQ.

\subsection{Concept Representations from Pairwise-link LDA}
\label{sec:lda}
\begin{figure}[!h]
  \begin{center}
\begin{tikzpicture}[scale=0.4]
  
  \node[obs]                   (wd)      {$w_{dn}$} ; %
  \node[latent, above=0.4cm of wd]    (zd) {$z_{dn}$} ; %
  \node[latent, above=0.5cm of zd]    (theta_d)  {$\theta_d$}; %
  
  \node[obs, right=2.2cm of zd, yshift=-0.3cm]   (p)      {$e_{dd'}$} ; %
  \node[latent, left=0.5cm of p]    (zdd_) {$z_{dd'}$};
  \node[latent, right=0.5cm of p] (zd_d) {$z_{d'd}$};
  \node[latent, above=0.5cm of p] (eta) {$\eta$};
  
  \node[obs, right=5.50cm of wd]  (wd_)      {$w_{d'n}$} ; %
  \node[latent, above=0.4cm of wd_]    (zd_) {$z_{d'n}$} ; %
  \node[latent, above=0.5cm of zd_]    (theta_d_)  {$\theta_{d'}$}; %

  \node[const, above=0.2cm of eta] (atheta) {$\alpha_\theta$};
  \node[latent, below=1.0cm of p] (beta) {$\beta$};

  \plate [inner sep=5pt] {plate1} { 
    (wd)
    (zd)
  }  {$\forall n \in N_d$}; 
  \plate  [inner sep=5pt] {} { 
    (plate1) 
    (theta_d)
  } {$\forall d \in \mathcal{D}$} ;

  \edge {atheta} {theta_d} 
  \edge {theta_d} {zd}
  \edge {zd} {wd}
  \edge {beta} {wd}
  
  \plate [inner sep=5pt] {plate2} { 
    (wd_)
    (zd_)
  }  {$\forall n \in N_{d'}$}; 
  \plate [inner sep=5pt] {} { 
    (plate2) 
    (theta_d_)
  } {$\forall d' \in \mathcal{D}$} ; 
 
  \edge {atheta} {theta_d_} 
  \edge {theta_d_} {zd_}
  \edge {zd_} {wd_}
  \edge {beta} {wd_}
 
  \plate {} { 
    (zdd_) (p) (zd_d) 
  } {$\forall (d,d') \in \mathcal{D \times D}$}; 
  
  \edge {theta_d} {zdd_}
  \edge {theta_d_} {zd_d}
  \edge {zdd_} {p}
  \edge {zd_d} {p}
  \edge {eta} {p}
  
\end{tikzpicture}
  \end{center}
  \caption{\small Graphical representation of the Pairwise Link-LDA model. Topicality of the document is explicitly made dependent on prerequisite documents using the variable $e_{dd'} \in E_D$, the observed prerequisite relation between the documents.}
  \label{fig:lda_model}
\end{figure}
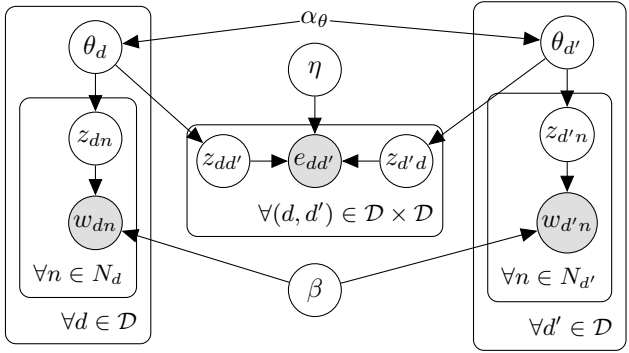
The Pairwise-Link-LDA model \cite{nallapati2008joint}combines the ideas of Latent Dirichlet Allocation (LDA) \cite{blei2003latent} and Mixed Membership Block Stochastic Models \cite{airoldi2008mixed} for jointly modeling text and links between documents in the topic modeling framework. Document graph $G_D$ and the set of text documents $\mathcal{D}$ are input to this model. 
A mixed membership model is a natural choice for modeling the documents, each of which includes many key concepts from different underlying topics. Figure \ref{fig:lda_model} shows the graphical representation of the generative model.  
Explicit modeling of the directional links (i.e., prerequisite edges) between an ordered pair of documents ($E_D$) captures the topicality of documents and the word distribution over topics better, in terms of capturing the prerequisite relationship between the words itself.

\subsubsection{Generative Process and Inference.}

Figure \ref{fig:lda_model} shows that
each document generation process is same as in LDA. 
Each text unit ($w_{dn}$), in our case an $n$-gram,  is generated from a topic ($z_{dn}$) sampled from 
the document-topic distribution ($\theta_d$).
The topic-word distribution $\beta$ describes the topic distribution of each word.
For each pair of documents $(d,d')$, the observed Bernoulli random variable $e_{d,d'}$ denotes the presence or absence of a prerequisite link from $d$ to $d'$ (i.e. an edge in $E_D$). 
For each document $d \in \mathcal{D}$, $z_{dn}$ is the index of the topic that generates the $n$th unit $w_{dn}$ in document $d$. For each pair of documents $(d,d')$, the latent topic sampled from $d$ for the prerequisite is $z_{dd'}$ and similarly, the latent topic sampled from $d'$ is $z_{d'd}$. 
The topic $z_{dd'}$ is sampled from the same document-word distribution ($\theta_d$) that is used to generate the document, thus modeling the dependence of the prerequisites on the underlying topics. 

It is important to note that the asymmetric prerequisite relation between pairs of documents is modeled by a Bernoulli random variable whose parameter is dependent on the underlying topics in the document pair.
The Bernoulli parameter $\eta_{z_{dd'}, z_{d'd}}$ enables asymmetric directionality in the prerequisite link.
For example, let $d$ be a prerequisite of $d'$ and $z_{dd'}$ and $z_{d'd}$ be the latent topics sampled from $d$ and $d'$ respectively for this interaction.
Then the parameter used to generate the Bernoulli random variable $e_{d,d'}$ will be $\eta_{z_{dd'}, z_{d'd}}$ which is different from $\eta_{z_{d'd}, z_{dd'}}$ thus modeling the directionality in the relationship.

We refer the reader to \cite{nallapati2008joint} for more details of the model and a mean--field variational approximation to infer the model parameters. Using this approach, we train the model with our input document corpus and their corresponding prerequisite links, with a fixed value of the hyperparameter $\alpha$. We learn $\beta_{K \times |V|}$, the word distribution over topics and $\eta_{K \times K}$, the asymmetric relationship between each pair of topics, where $V$ is our vocabulary of $n$-grams and $K$ is the chosen number of topics.

The inferred $\eta$ matrix shows the asymmetric pairwise relation between underlying topics in the document corpus.
A natural approach to learning prerequisite relations is to use the inferred topic distribution for each concept word (from $\beta$, after suitable normalization) and topic prerequisite relation $\eta$ to learn 
concept prerequisiteness. 
However, our experiments show that this approach does not work well.

\subsection{Predicting Relations using Siamese Network}
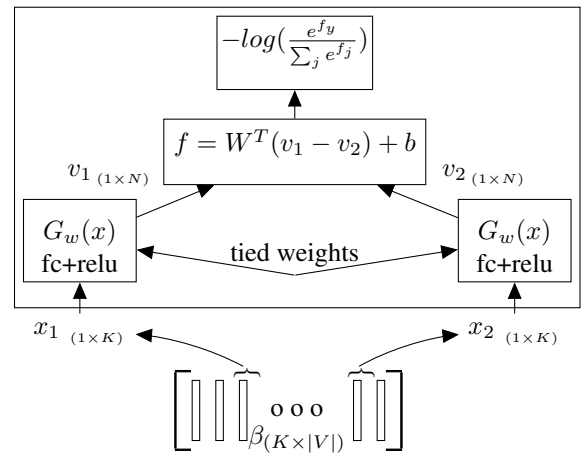
\begin{figure}[!h]
  \begin{center}
    \tikzstyle{plate} = [draw, rectangle,  fit=#1]
\tikzstyle{vec} = [rectangle, fill=white,draw=black, inner sep=0pt, node distance=1, minimum width=0.1cm, minimum height = 0.8cm,  font=\fontsize{10}{10}\selectfont]

\tikzstyle{hvec} = [rectangle, fill=white,draw=black, inner sep=2pt, node distance=1, minimum width=2cm,  minimum height = 0.5cm,  font=\fontsize{10}{10}\selectfont]

\tikzstyle{vtext} = [label, distance=0.5cm,text depth=-1ex,rotate=-90]
\tikzstyle{vedge} = [rectangle, fill=black,draw=black, inner sep=0pt, node distance=1, minimum width=0.5pt,  minimum height =1.0cm]
\tikzstyle{hedge} = [rectangle, fill=black,draw=black, inner sep=0pt, node distance=1, minimum width=0.2cm,  minimum height =0.5pt]

\tikzstyle{empty} = [rectangle, fill=white,draw=white, inner sep=0pt, node distance=1, minimum width=0.0cm, minimum height = 0.0cm,  font=\fontsize{10}{10}\selectfont]

\begin{tikzpicture}[scale=0.5]

  
  \node[label]                   	(x1)      {$x_{1_{ \;\;(1 \times K)}}$} ; %
  \node[label, right=4.3cm of x1] 	(x2)      {$x_{2_{ \;\;(1 \times K)}}$} ; %
  
  \node[const, above=0.9cm of x1]  	(g1) {$G_w(x)$};
  \node[const, below=0.1cm of g1]  	(l1) {fc+relu};
    \plate [inner sep=7pt] {plate1} { 
    (g1)
  }  {};

  \node[const, above=0.9cm of x2]  	(g2) {$G_w(x)$};
  \node[const, below=0.1cm of g2]  	(l2) {fc+relu};
  \plate [inner sep=7pt] {plate2} { 
    (g2)
  }  {};

  \node[label, above=0.3cm of g1, xshift=0.4cm] (o1)  {$v_{1_{\;(1 \times N)}}$} ;
  \node[label, above=0.3cm of g2, xshift=-0.4cm] (o2) {$v_{2_{\;(1 \times N)}}$} ;

  \node[const, right=1.5cm of g1, yshift=-0.3cm] 	(w)      {tied weights} ;
  \node[empty, below=0.1cm of w]	(platew)	{};

  \node[label, above=1.0cm of w]	(f)		{$f=W^T(v_1-v_2)+b$};
  \plate [inner sep=0pt] {platef} { 
    (f)
  } {};
  \node[const, above=2.3cm of w] 	(loss)      {$- log(\frac{e^{f_{y}}}{\sum_j e^{f_j}})$ } ;
  \plate [inner sep=1pt] {plateloss} { 
    (loss)
  }  {};
  
  \plate  [inner sep=3pt] {} { 
     (plate1) (platew) (plate2)
     (plateloss)
  } {} ;

  \node[vec, below=1.5cm of w, xshift=-1.0cm] (v1) {};
  \node[vec, right= 0.2cm of v1] (v2) {};
  \node[vec, left= 0.2cm of v1] (v3) {};
  \node[const, right= 0.3cm of v2] (dot) {o o o};
  \node[vec, right= 0.7cm of dot] (v4) {};
  \node[vec, left= 0.2cm of v4] (v5) {};
  \node[const, below= 0.1cm of dot] (beta) {$\beta_{(K \times |V|)}$};

  \node[vedge, left= 0.2cm of v3] (lb) {};
  \node[vedge, right= 0.2cm of v4] (rb) {};
  \node[hedge, below= 0.0cm of lb, xshift=0.1cm] (lbb) {};
  \node[hedge, above= 0.0cm of lb, xshift=0.1cm] (lbt) {};
  \node[hedge, below= 0.0cm of rb, xshift=-0.1cm] (rbb) {};
  \node[hedge, above= 0.0cm of rb, xshift=-0.1cm] (rbr) {};

  \node[vtext, above= 0.1cm of v2] (brace1) {\{};
  \node[vtext, above= 0.1cm of v5] (b2) {\{};
  \node[label, above= 0cm of v5, xshift=-0.1cm] (brace2) {};

  \edge {x1} {plate1}
  \edge {x2} {plate2}
  \edge {plate1} {platef}
  \edge {plate2} {platef}
  \edge {platef} {plateloss}
  \edge {platew} {plate1}
  \edge {platew} {plate2}
  \path[->] (brace1) edge [bend right=12,looseness=0.8] (x1);
  \path[->] (brace2) edge [bend left=12,looseness=0.8] (x2);

\end{tikzpicture}
  \end{center}
  \caption{\small Siamese Architecture. Each branch consists of 2 fully connected (FC) layers with ReLU nonlinearities between them with tied weights. We train this network with the concept vectors learned from Pairwise link-LDA model (with $K$ topics and vocabulary $V$), using cross-entropy loss, with positive and negative concept pairs, $y \in \{0,1\}$ is the label of the corresponding pair.}
  \label{fig:siamese_model}
\end{figure}

A Siamese network generally comprises of two identical sub-networks that are joined by one cost module \cite{siamese}. 
The architecture is shown in Figure \ref{fig:siamese_model}.
Each input to our Siamese Network is a pair of vectors ($x_1,x_2$) and a binary label $\in \{0,1\}$. 
The weights of the sub-networks are tied and each subnetwork is denoted as $G_w(.)$.
The pair ($x_1,x_2$) is passed through the sub-networks ($G_w$) of two fully connected (FC) neural network layers and a rectified linear unit (ReLU), yielding two corresponding outputs ($v_1,v_2$).
The loss function is optimized with respect to the parameter vectors controlling both the subnets through stochastic gradient decent method using the Adam optimizer.

\section{PREREQ: Concept Prerequisite Prediction}

We use exponentiated and normalized columns of $\beta$ as vector representations of the concepts.
Hence, each concept is represented as a $[1 \times K]$ dimensional vector where $K$ is the number of latent topics. Labeled pairs of such vectors from our training set $E_{CT}$ are used as input to train the Siamese network.
The label is set to 1 when the first concept is a prerequisite to the second and 0, otherwise.
These vectors are passed through the sub-networks, $G_w(.)$. We use the sum of the weighted element-wise differences between the twin feature vectors $v_1$ and $v_2$. $W$ and $b$ denotes the weights and biases that connects the difference of the outputs from the two sub-networks to the loss layer. 
We use the cross-entropy loss function and obtain the probability ($P_{c_s, c_t}$) of the first input vector ($c_s$) being a prerequisite to the second ($c_t$): 
\begin{equation}
P_{c_s, c_t} = log\left(\frac{e^{f_{y_i}}}{\sum_{j\in\{0,1\}} e^{f_j}}\right)
\label{eqn:score}
\end{equation}
where $f_j$ is the $j^{th}$ element of the 2-dimensional vector \textbf{$f = W^T(G_w(\beta^{T}_{c_s}) - G_w(\beta^{T}_{c_t})) + b$}, and $y_i=1$, as we are solving a binary classification problem. 
The trained Siamese network can be used for predicting prerequisite relations.

\section{Related Work}
Inferring concept prerequisites, from course dependencies or from video based course data are relatively new areas of study.
To our knowledge, there have been three previous methods specifically designed to infer educational concept prerequisites, viz. 
CGL \cite{liu2016learning},
CPR-Recover \cite{eaai17}, and
MOOC-RF \cite{acl2017prerequisite}.
{\bf CGL} is a supervised learning approach to map courses from different universities onto a universal space of concepts and predict prerequisites between both courses and concepts \cite{liu2016learning}.
They represent the courses in vector space and use ranking and another classification based approach. 
{\bf CPR-Recover}
solves the same problem by
formulating a quadratic optimization problem \cite{eaai17} and shows better performance than CGL. 
But the number of constraints in the optimization problem is proportional to the number of course prerequisite edges, which does not scale well with the size of the training data.
Pan {\it et al.} recently proposed a method {\bf MOOC-RF} for concept prerequisite recovery from Coursera data \cite{acl2017prerequisite}. 
They define various features 
and train a classifier that can identify prerequisite relations among concepts from video transcripts.
They do not use course/video prerequisite pairs to infer concept prerequisites.
Both CGL and MOOC-RF 
use semantics and context based features.
Our method, instead of using hand tuned features,
utilizes a pairwise generative model to automatically learn the features from the hidden representation in order to infer concept prerequisite edges.

\section{Experiments}

We first compare the performance of PREREQ (source code{\footnote{{https://github.com/suderoy/PREREQ-IAAI-19/}}}) with that of other state-of-the-art algorithms for inferring prerequisite relations on benchmark datasets.
We consider both cases of course prerequisites as well as video playlists, as input, with corresponding baseline methods.
Additionally we demonstrate how PREREQ can learn effectively even when there is less training data available.

\subsection{Data}
We use a published benchmark dataset, the University Course Dataset and in addition create a new dataset as described below. Dataset statistics are detailed in table \ref{tab:datasets}.

\begin{table}[h]
  \centering
  \small
  \begin{tabular}{|l|c|c|c|c|}\hline
  Dataset & $|\mathcal{D}|$ & $|E_D|$ & $|E_C|$ & $|C|$\\ \hline
  University Course Dataset & 654 & 861 & 1008 & 365\\
  NPTEL MOOC Dataset & 382 & 1445 & 1008 & 345\\
      \hline
  \end{tabular}
  \caption{\small Dataset Statistics.}
    \label{tab:datasets}
\end{table}

\noindent
\textbf{University Course Dataset.}
This dataset, from \cite{eaai17}, has 654 courses, from various universities in USA, and 861 course prerequisite edges. Manual annotation of 1008 pairs of concepts with prerequisite relations are provided. There are 406 unique concepts (word or phrases) among which 1008 prerequisite relationships are annotated.\\
\textit{Data Preparation.} 
We create bag-of-words (BoW) for unigram, bigram and trigrams from each course text, removing the standard English stopwords, to represent each course as a BoW vector. The BoW vectors of 654 courses and 861 course prerequisite edges are used by the pairwise link-LDA model to infer  concept vectors. We lemmatize the given concepts to match with the concepts from the BoW vocabulary and get 365 concepts from the vocabulary out of the 406. The concept vectors, which are inferred by the pairwise link-LDA, and the 1008 concept prerequisite pairs are used by the Siamese network for 5 train-test splits as described below.\\

\noindent
\textbf{MOOC Dataset.}
This dataset is based on video playlists from a MOOC corpus.
We download the subtitles of the videos from playlists of computer science departments from NPTEL {\footnote{{http://nptel.ac.in/}}}. We use 382 videos from 38 different playlists. The same 1008 concept prerequisite pairs from the University Course Dataset are used as annotated concept pairs, since both the datasets are based on computer science courses. \\
\textit{Data Preparation.} 
We use the video subtitle text (i.e. speech transcripts) to create the BoW vectors of the videos and the words and phrases present in concept vocabulary are considered as the vocabulary for creating BoW. Using similar pre-processing as for the previous dataset, here we find 345 concepts from the BoW vocabulary. As there is no video-video (or course) prerequisite edge present in this case, we use the temporal relatedness as a proxy for course prerequisite relationships. That is, a video lecture in a particular playlist is a prerequisite for all the videos that are in the same playlist after the particular video, which
may add some noise in the form of erroneous edges.
This gives us total 1455 prerequisite edges between video pairs.

\subsection{Performance Evaluation}
\label{expts} 

To evaluate the performance of PREREQ on the datasets, we split the concept prerequisite edges into train and test sets. 60\% of the given concept pairs from the datasets are used for training while the rest 40\% for testing. 
Training a binary classifier requires both positive and negative  instances.
But the University Course dataset has only positive samples: the concept prerequisite pairs that are manually annotated. We generate negative samples by sampling random unrelated pairs of phrases from the vocabulary in addition to the reverse pair of original positive samples, to enable our model to learn prerequisite directionality. We oversample the negative instances to 1.5 times the number of positive examples in the training set to address the imbalance. All presented results are averaged over 5 train-test splits.

\subsubsection{PREREQ Parameter Settings.}

For all our experiments, we choose number of topics $K$ = 100 and a fixed Dirichlet parameter $\alpha$ = 0.01, to enable sparse topic distribution. 
The Siamese network is trained with learning rate of 0.0001 and batch size of 128 over 3500 iterations. 

\subsubsection*{Evaluation Metric.} 
To compare the performance of PREREQ,
we use the same evaluation metric used 
in our main baseline CPR-Recover \cite{eaai17}, i.e., Precision@K = $\frac{\sum_{i=1}^K rel(i)} {K}$, where rel(.) is a binary indicator of presence of the concept pair $(c_s,c_t)$ in the ground truth. We sort all concept pairs based on their probability (as in Eq. \ref{eqn:score} predicted by PREREQ) and choose top K = 50 or K = 100 to calculate the precision. The x-axis in the performance graphs denotes the number of course prerequisite edges used to predict the concept prerequisite pairs. In addition we also use Precision, Recall and F-score to evaluate performance of PREREQ with all the baselines.

\subsubsection{Baseline Methods.} 

The method \textbf{CPR-Recover} \cite{eaai17} is, to our knowledge, the best known method (though unsupervised) for inferring concept prerequisites as tested on the University Course Dataset, where they have shown that it outperforms previous methods including CGL \cite{liu2016learning}. 
\textbf{MOOC-RF} \cite{acl2017prerequisite} is a supervised method, designed for online video based courses. Note that they have a different problem setting and do not use course/video prerequisite pairs to infer concept prerequisites. Hence, precision@Kfor using different sets of course prerequisite edges is not a computable measure for this method. So, we compare with MOOC-RF using metrics precision, recall and F-score. 

In addition to these baselines, we  use a simple count-based method, \textbf{Freq}, that calculates the score of a concept pair based on the number of times the pair `co-occurs' in course prerequisite pairs as described in \cite{eaai17}.
Also, based on the parameters ($\beta$ and $\eta$) inferred from the pairwise link-LDA model, we predict the directed relationship between a pair of concepts $c_s$ and $c_t$ by computing the score 
$ s_{c_s,c_t} = \beta_{c_s}^T \eta \beta_{c_t}$. Each column of $\beta$ is exponentiated and normalized by dividing all element of the column by the maximum element. For precision, recall and F-score computation we use a threshold of 0.5 on the score to distinguish the classes. We call this method \textbf{Pairwise LDA}. 
\subsection{Performance on Benchmark Datasets}

\begin{figure} [!h]
\begin{minipage}{0.49\linewidth}
 \centering
 \includegraphics[width=\linewidth]{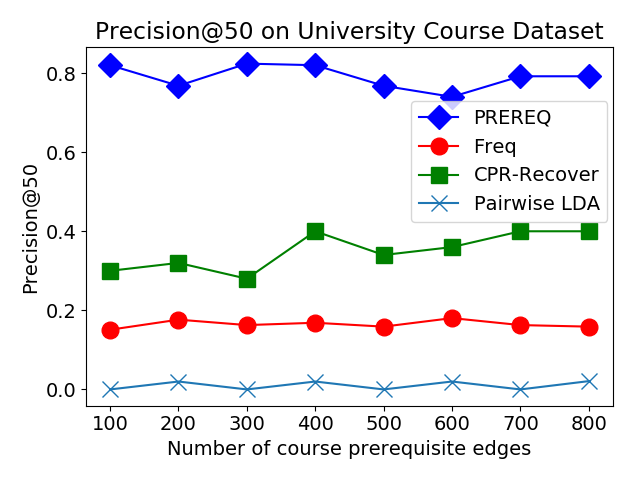}
\end{minipage}
\begin{minipage}{0.49\linewidth}
\centering
 \includegraphics[width=\linewidth]{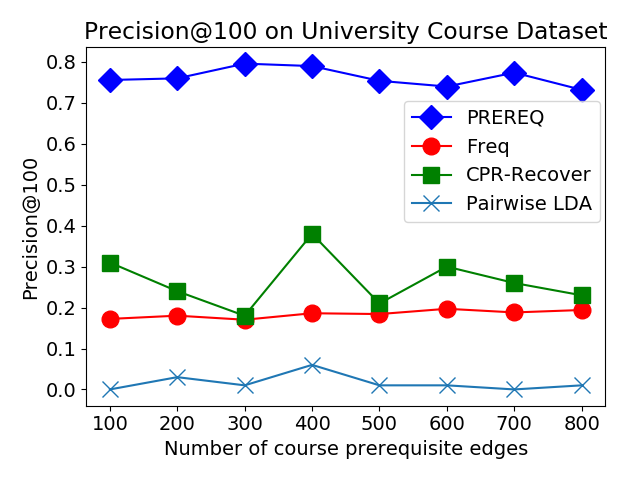}
\end{minipage}
 \caption{\small PREREQ shows significant improvement in Precision@50 and Precision@100, on the University Course Dataset. The performance is consistently better even when lesser course prerequisites ($E_D$) are used.}
\label{fig:eaai_P@k}
\end{figure}

\begin{figure}[!h]
\begin{minipage}{0.49\linewidth}
 \centering
 \includegraphics[width=\linewidth]{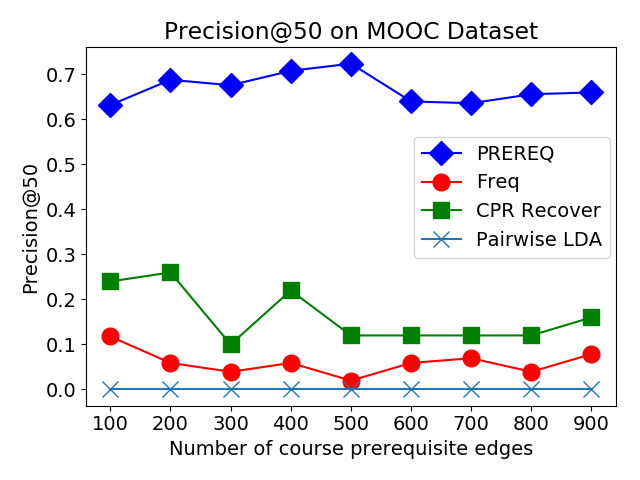}
\end{minipage}
\begin{minipage}{0.49\linewidth}
 \centering
 \includegraphics[width=\linewidth]{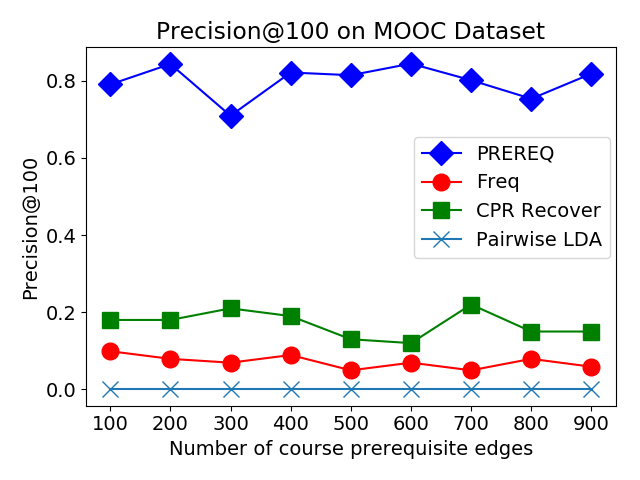}
\end{minipage}
 \caption{\small Results on the MOOC Dataset. PREREQ accurately retrieves the concept prerequisite edges with high probability, as measured by Precision@50 and Precision@100, even on video playlist data where course prerequisite links are unavailable.}
\label{fig:nptel_P@k}
\end{figure}

From the University Course Dataset, {100, 200, ..., 800} course prerequisite edges are randomly sampled and precision@K values are computed and averaged over multiple iterations. 
Figure \ref{fig:eaai_P@k} shows the comparative results where PREREQ performs significantly better than CPR-Recover consistently over different number of tested prerequisite edges. 
Figure \ref{fig:nptel_P@k} shows the Precision@K scores for K = 50 and K =100 on the MOOC dataset obtained from the NPTEL video playlists. The results demonstrate that on video playlists, where course/video prerequisite information is {\it not} available, PREREQ is able to infer concept prerequisites accurately.

\begin{figure}[!h]
 \begin{minipage}{0.49\linewidth}
 \centering
 \includegraphics[width=1\linewidth]{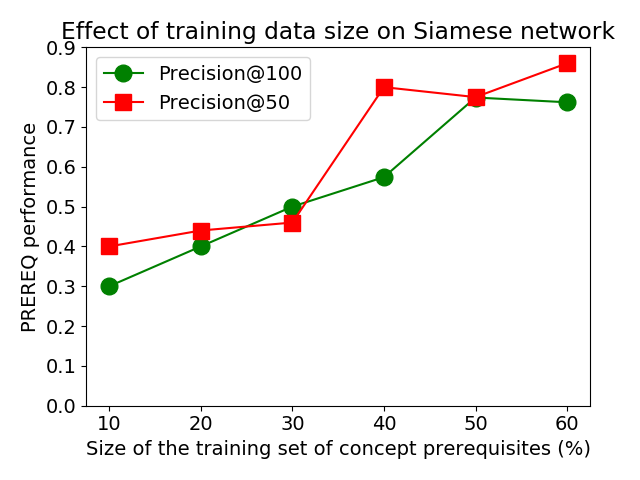}
 {\small (a) Performance of PREREQ}
 \end{minipage}
 \begin{minipage}{0.5\linewidth}
 \centering
 \includegraphics[width=1\linewidth]{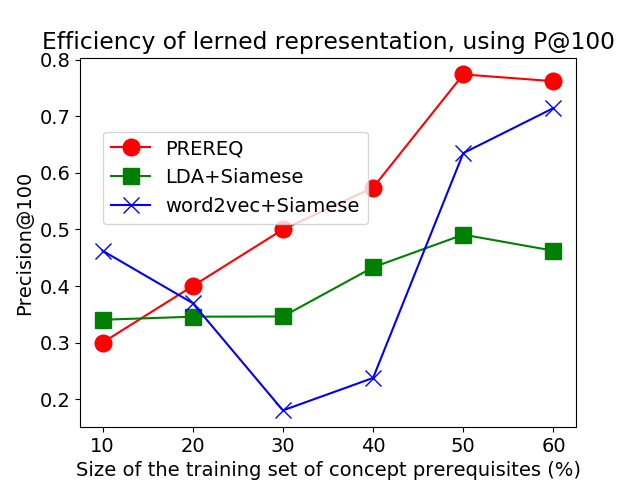}
 {\small (b) Representation comparison}
 \end{minipage}
 \caption{\small Results on University Course dataset, (a) Effect of Training Data Size (b) Effect of Concept Representations.}
\label{fig:train_test}
\end{figure}

\begin{table*}
  \centering
  \footnotesize
  \begin{tabular}{|l|cccc|cccc|}\hline
	  Dataset & \multicolumn{4}{c|}{University Course Dataset} & \multicolumn{4}{c|}{ MOOC Dataset } \\
      \hline
    Method & PREREQ &Pairwise LDA&CPR-Recover &MOOC-RF& PREREQ &Pairwise LDA&CPR-Recover &MOOC-RF\\
    \hline
    Precision & 46.76 &\textbf{98.27} & 16.66 & 43.70 & 55.60 &48.43& 17.18 & \textbf{59.74}\\
    Recall & \textbf{91.64} &16.42& 46.51 & 53.43 & \textbf{75.74} &10.47& 52.97 & 56.48\\
    F-score & \textbf{59.68} &28.14& 24.54 & 50.95& \textbf{60.73} &17.22& 25.94 & 58.07\\
    \hline
  \end{tabular}
  \caption{\small Performance of PREREQ on benchmark the University Course Dataset and MOOC dataset. Row-wise best results in bold.}
  \label{tab:PR_eaai}
\end{table*}
Table \ref{tab:PR_eaai} shows the precision, recall and F-score on both the University Course Dataset and MOOC Dataset. The F-score of PREREQ is higher than that of CPR-Recover and MOOC-RF in both the datasets.

\subsubsection{Effect of Training Data Size on Performance.}
Labeled concept prerequisites may be hard to obtain and may not be sufficient to train a supervised model. 
However the simple Siamese architecture in PREREQ uses very few parameters and can be trained easily even with less training data as seen in Figure \ref{fig:train_test}(a) where the  performance of PREREQ on different amounts of training data is compared. We find that the performance reduces only marginally even when only 40\% of the available labeled data is used for training.

\subsubsection{Effect of Concept Representations.}
Our experiments (not shown) show that inferred topics from pairwise-link LDA can discriminate between related and unrelated documents (based on the prerequisite relation), but the topics do not have sufficient signal to determine the directionality of the prerequisite edge. Nevertheless, inferred topics are good concept representations. We test the performance of three different concept representations -- (1) topic distributions obtained from pairwise-link LDA (as used in PREREQ), (2) topic distributions obtained directly from LDA learnt from the same document corpus, and (3) word2vec representations \cite{w2v}, trained over Wikipedia. Figure \ref{fig:train_test}(b) shows that pairwise-link LDA based concept representations are superior to those based on LDA and word2vec.

\section{Conclusion}
We develop PREREQ, a supervised learning method, to learn concept prerequisites from course prerequisite data and from unlabeled video playlists (increasingly available from MOOCs), that obviates the need for manual creation of labeled course prerequisite datasets.
PREREQ obtains latent representations of concepts through the pairwise-link LDA model, which are then used to train a Siamese network that can identify prerequisite relations accurately. 
PREREQ outperforms state-of-the-art methods for inferring prerequisite relations between educational concepts, on benchmark datasets. 
We also empirically show that PREREQ can learn effectively from very less training data and from unlabeled video playlists. 
PREREQ can effectively utilize the large and increasing amount of online educational material in the form of text (course webpages) and video (MOOCs) to solve a fundamental problem that is essential for several online educational technology applications.

\bibliographystyle{aaai}
\bibliography{biblio}  

\begin{thebibliography}{}

\bibitem[\protect\citeauthoryear{Airoldi \bgroup et al\mbox.\egroup
  }{2008}]{airoldi2008mixed}
Airoldi, E.~M.; Blei, D.~M.; Fienberg, S.~E.; and Xing, E.~P.
\newblock 2008.
\newblock Mixed membership stochastic blockmodels.
\newblock {\em Journal of Machine Learning Research} 9(Sep):1981--2014.

\bibitem[\protect\citeauthoryear{Bengio, Courville, and
  Vincent}{2013}]{bengio2013representation}
Bengio, Y.; Courville, A.; and Vincent, P.
\newblock 2013.
\newblock Representation learning: A review and new perspectives.
\newblock {\em IEEE Transactions on Pattern Analysis and Machine Intelligence}
  35(8):1798--1828.

\bibitem[\protect\citeauthoryear{Blei, Ng, and Jordan}{2003}]{blei2003latent}
Blei, D.~M.; Ng, A.~Y.; and Jordan, M.~I.
\newblock 2003.
\newblock Latent dirichlet allocation.
\newblock {\em Journal of Machine Learning Research} 3(Jan):993--1022.

\bibitem[\protect\citeauthoryear{Bromley \bgroup et al\mbox.\egroup
  }{1994}]{siamese}
Bromley, J.; Guyon, I.; LeCun, Y.; S{\"a}ckinger, E.; and Shah, R.
\newblock 1994.
\newblock Signature verification using a siamese time delay neural network.
\newblock In {\em NIPS}.

\bibitem[\protect\citeauthoryear{Jardine}{2014}]{jardine2014automatically}
Jardine, J.~G.
\newblock 2014.
\newblock Automatically generating reading lists.
\newblock Technical report, University of Cambridge.

\bibitem[\protect\citeauthoryear{Laurence and
  Margolis}{1999}]{laurence1999concepts}
Laurence, S., and Margolis, E.
\newblock 1999.
\newblock Concepts and cognitive science.
\newblock {\em Concepts: Core Readings}  3--81.

\bibitem[\protect\citeauthoryear{Liang \bgroup et al\mbox.\egroup
  }{2017}]{eaai17}
Liang, C.; Ye, J.; Wu, Z.; Pursel, B.; and Giles, C.~L.
\newblock 2017.
\newblock Recovering concept prerequisite relations from university course
  dependencies.
\newblock In {\em AAAI}.

\bibitem[\protect\citeauthoryear{Liang \bgroup et al\mbox.\egroup
  }{2018}]{liang2018investigating}
Liang, C.; Ye, J.; Wang, S.; Pursel, B.; and Giles, C.~L.
\newblock 2018.
\newblock Investigating active learning for concept prerequisite learning.
\newblock In {\em EAAI}.

\bibitem[\protect\citeauthoryear{Liu \bgroup et al\mbox.\egroup
  }{2016}]{liu2016learning}
Liu, H.; Ma, W.; Yang, Y.; and Carbonell, J.
\newblock 2016.
\newblock Learning concept graphs from online educational data.
\newblock {\em Journal of Artificial Intelligence Research} 55:1059--1090.

\bibitem[\protect\citeauthoryear{Mikolov \bgroup et al\mbox.\egroup
  }{2013}]{w2v}
Mikolov, T.; Sutskever, I.; Chen, K.; Corrado, G.~S.; and Dean, J.
\newblock 2013.
\newblock Distributed representations of words and phrases and their
  compositionality.
\newblock In {\em NIPS}.

\bibitem[\protect\citeauthoryear{Nallapati \bgroup et al\mbox.\egroup
  }{2008}]{nallapati2008joint}
Nallapati, R.~M.; Ahmed, A.; Xing, E.~P.; and Cohen, W.~W.
\newblock 2008.
\newblock Joint latent topic models for text and citations.
\newblock In {\em ACM SIGKDD}.

\bibitem[\protect\citeauthoryear{Novak}{1990}]{novak1990concept}
Novak, J.~D.
\newblock 1990.
\newblock Concept mapping: A useful tool for science education.
\newblock {\em Journal of Research in Science Teaching} 27(10):937--949.

\bibitem[\protect\citeauthoryear{Pan \bgroup et al\mbox.\egroup
  }{2017}]{acl2017prerequisite}
Pan, L.; Li, C.; Li, J.; and Tang, J.
\newblock 2017.
\newblock Prerequisite relation learning for concepts in moocs.
\newblock In {\em ACL}, volume~1,  1447--1456.

\bibitem[\protect\citeauthoryear{Rouly, Rangwala, and
  Johri}{2015}]{rouly2015we}
Rouly, J.~M.; Rangwala, H.; and Johri, A.
\newblock 2015.
\newblock What are we teaching?: Automated evaluation of cs curricula content
  using topic modeling.
\newblock In {\em Proceedings of the 11th Annual International ACM Conference
  on International Computing Education Research},  189--197.

\bibitem[\protect\citeauthoryear{Talukdar and Cohen}{2012}]{Talukdar:2012}
Talukdar, P.~P., and Cohen, W.~W.
\newblock 2012.
\newblock Crowdsourced comprehension: Predicting prerequisite structure in
  wikipedia.
\newblock In {\em 7th Workshop on Building Educational Applications Using NLP}.

\bibitem[\protect\citeauthoryear{Yosef \bgroup et al\mbox.\egroup
  }{2011}]{yosef2011aida}
Yosef, M.~A.; Hoffart, J.; Bordino, I.; Spaniol, M.; and Weikum, G.
\newblock 2011.
\newblock {AIDA}: An online tool for accurate disambiguation of named entities
  in text and tables.
\newblock {\em Proceedings of the VLDB Endowment} 4(12):1450--1453.

\end{thebibliography}

\end{document}